\documentclass[a4paper]{article}

\usepackage{INTERSPEECH2020}
\usepackage{hyperref}
\usepackage{breakurl}
\usepackage[group-separator={,}]{siunitx}
\usepackage{tikz}
\usepackage{pgfplots}
\usepgfplotslibrary{groupplots}
\usepackage{multirow}
\usepackage{flushend}

\usepackage{xcolor}
\usepackage[colorinlistoftodos,prependcaption,textsize=footnotesize]{todonotes}
\newcommand{\librispeech}{\textsc{\mbox{LibriSpeech}} dataset}

\newcommand{\librilight}{\textsc{\mbox{LibriLight}} dataset}

\def\Snospace~{\S{}}

\def\devother{\texttt{dev-other}}

\makeatletter
\newcommand\notsotiny{\@setfontsize\notsotiny\@vipt\@viipt}
\makeatother

\newcommand*{\metric}[1]{\num[round-mode=places,round-precision=1]{#1}}

\title{Self-Training for End-to-End Speech Translation}
\name{Juan Pino$^1$, Qiantong Xu$^1$, Xutai Ma$^{1,2}$, Mohammad Javad Dousti$^1$, Yun Tang$^1$}
\address{
	$^1$Facebook AI, USA\\
	$^2$Johns Hopkins University, USA}
\email{\{juancarabina,qiantong,dousti,yuntang\}@fb.com, xutai\_\thinspace ma@jhu.edu}

\begin{document}
	
	\maketitle
	\begin{abstract}
		One of the main challenges for end-to-end speech translation is data scarcity. We leverage pseudo-labels generated from unlabeled audio by
		a cascade and an end-to-end speech translation model.
		This provides 8.3 and 5.7 BLEU gains over a strong semi-supervised baseline on the MuST-C English-French and English-German datasets, reaching state-of-the art performance.
		The effect of the quality of the pseudo-labels is investigated.
		Our approach is shown to be more effective than simply
		pre-training the encoder on the speech recognition task. Finally,
		we demonstrate the effectiveness of self-training by directly generating pseudo-labels with an end-to-end model instead of a cascade model.
	\end{abstract}
	
	\section{Introduction}
	
	Speech translation (ST) systems convert input audio in a language into text translations in another language. Compared with their cascade counterpart, end-to-end models have lower inference latency, are smaller and are less susceptible to error compounding. However, their main disadvantage comes from the lack of supervised training data.
	
	Data scarcity has been addressed in previous work with data augmentation~\cite{jia2019leveraging, pino2019harnessing}, multi-task training~\cite{Weiss2017, berard2018end},  pre-training~\cite{bansal-etal-2019-pre, 9053847} or multilingual speech translation~\cite{9004003, 9003832, wang2020covost}. In this paper, we propose to revisit self-training~\cite{scudder1965probability} in the context of speech translation. Labels are automatically generated from unlabeled audio data either via a strong speech recognition (ASR) system followed by a strong machine translation (MT) system, i.e.\ a cascade model, or via an end-to-end model. An end-to-end speech translation model is then trained on the resulting data.
	
	\cite{He2020Revisiting} demonstrates the effectiveness of self-training for machine translation and summarization. They also provide insights into its success and further improve vanilla self-training by introducing noise in the unlabeled data.
	\cite{synnaeve2019end} and \cite{kahn2019self} also leverage pseudo-labeling on the \librilight~\cite{librilight} to improve the performance of an end-to-end ASR system. Similar to this work, additional knowledge (i.e., additional monolingual data to train the language model) is leveraged to generate the pseudo-labels.
	\cite{yalniz2019billion} also explores pseudo-labeling (both knowledge distillation and self-training) at scale in the domain of computer vision.
	\cite{jia2019leveraging} demonstrates how to improve the performance of an end-to-end speech translation system by generating pseudo-labels from unlabeled audio via a cascade system. In contrast, we provide more insights on the conditions under which this method works. Furthermore, we demonstrate how to generate pseudo-labels with an end-to-end system, which may simplify model building. Finally, we conduct experimentation on open benchmarks for reproducibility. \cite{9054759} also leverages additional ASR and MT resources to improve end-to-end speech translation with a meta-learning algorithm. Our approach aims at simplifying model building by reusing either off-the-shelf ASR and MT systems or an end-to-end speech translation for pseudo-labeling and obtains state-of-the-art results.
	
	Our method is first shown to be effective in a low resource setting (\autoref{sec:low_resource}). On a higher resource setup, improvements are obtained after fine-tuning on the baseline data and by training larger models (\autoref{sec:finetuning_transformer}). Since pseudo-labeling enables the training of larger
	architectures, scaling up the size of ST models is investigated next (\autoref{sec:scaling_model_size}). By doing so, we obtain large improvements over a strong semi-supervised baseline across three language pairs and two domains and reach state-of-the-art performance on the MuST-C English-French and English-German datasets.
	In ablation studies, our method is shown to be more effective than pre-training the encoder on the ASR task (\autoref{sec:selftraining_vs_pretraining}). We also study the effect of the quality of the pseudo-labels on the low resource setting (\autoref{sec:quality}). Finally, replacing the cascade model with an end-to-end model for pseudo-labeling is investigated (\autoref{sec:selftraining}).
	
	
	\section{Experimental Setup}
	
	\begin{table}[t]
		\centering
		\caption{Open and FB Video dataset statistics, reported after filtering for too long or too short input.}
		\resizebox{\columnwidth}{!}{%
			\begin{tabular}{lllrr}
				\toprule
				Domain & Language & Dataset         & \# utterances & \# hours \\
				\midrule
				\multirow{8}{*}{Open} & \multirow{3}{*}{En-Fr} & MuST-C & \num{275}k & \num{479} \\
				& & dev & \num{1412} & \num{2.6} \\
				& & tst-COMMON & \num{2632} & \num{4.2} \\
				\cmidrule(lr){2-5}
				& \multirow{3}{*}{En-De} & MuST-C & \num{230}k & \num{395} \\
				& & dev & \num{1423} & \num{2.5} \\
				& & tst-COMMON & \num{2641} & \num{4.1} \\
				\cmidrule(lr){2-5}
				& \multirow{2}{*}{En} & \textsc{LibriSpeech} & \num{281}k & \num{960} \\
				& & \textsc{Librilight} & \num{15.8}M & \num{56}k \\
				\midrule
				\multirow{8}{*}{FBVideos} & \multirow{3}{*}{En-Fr} & train & \num{20.7}M & \num{30}k \\
				& & dev & \num{925} & \num{6.3} \\
				& & test & \num{3909} & \num{24.3} \\
				\cmidrule(lr){2-5}
				& \multirow{3}{*}{En-Es} & train & \num{20.6}M & \num{30}k \\
				& & dev & \num{935} & \num{6.4} \\
				& & test & \num{3915} & \num{24.3} \\
				\cmidrule(lr){2-5}
				& En & unlabeled & \num{32.2}M & \num{255}k \\
				\bottomrule
			\end{tabular}
		}
		\label{tab:aststats}
	\end{table}
	
	\subsection{Data}
	
	Experiments are conducted with both open and proprietary data.
	Open data is used for reproducibility purposes and to conduct more detailed ablations while proprietary data, comprised of de-identified and aggregated public Facebook (FB) videos, is used to verify that our methods work at large scale. Open data includes the English-German and English-French portions of MuST-C~\cite{di-gangi-etal-2019-must}, \textsc{LibriSpeech}~\cite{panayotov2015librispeech} (LS) transcripts with automatic translations for a higher resource baseline and \textsc{LibriLight} (LL) to provide English unlabeled audio. Different amounts of English unlabeled data are randomly sampled and reused for all experiments. Three language pairs, English-German (En-De), English-French (En-Fr) and English-Spanish (En-Es) are studied. Dataset statistics are summarized in \autoref{tab:aststats}.
	%
	
	
	\subsection{Speech Translation Models}
	
	Models take log-mel filterbank features, computed with a \SI{10}{\milli\second} window shift, as input. On FB Video data, features have 40 dimensions and a window size of \SI{16}{\milli\second}, and utterances of more than 6000 frames are removed. On open data, features have 80 dimensions and a window size of \SI{25}{\milli\second}, and utterances with more than 4000 frames, less than 20 frames, or more than 256 tokens are removed. The translated text vocabulary is a unigram model with size 10,000 built with the SentencePiece~\cite{Kudo2018SentencePieceAS}. Note that a separate vocabulary is rebuilt for each data condition and that the model is directly built on raw data without pre-tokenization.
	
	We investigate our proposed method with a relatively small LSTM architecture and a large Transformer architecture~\cite{vaswani2017attention}. The LSTM architecture consists of a speech encoder with non-linear layers followed by convolutional layers and bidirectional LSTM layers, and a custom LSTM decoder~\cite{pino2019harnessing, berard2018end}. The Transformer architecture, \textsc{vggtransformer}, is an
	adaptation of Transformer to the ASR task~\cite{mohamed2019transformers}. Two architectures, \textsc{VggT}, with 14 encoder layers and 4 decoder layers, and \textsc{VggTLarge}, with 20 encoder layers and 10 decoder layers are used in experiments.
	
	Training uses the Adam optimizer~\cite{Kingma2015AdamAM} with a learning rate of 0.001 for the LSTM architecture and 0.0001 for the \textsc{vggtransformer} architecture. The LSTM architecture has a fixed learning rate schedule while the \textsc{vggtransformer} architecture uses the original Transformer learning rate schedule~\cite{vaswani2017attention}. Mini-batches have an effective size of 384,000 frames. Models are trained until convergence or up to 800k updates.
	
	At inference time, we use beam search with beam size 20, including for pseudo-label generation. Case-sensitive detokenized BLEU is computed with SacreBLEU~\cite{post-2018-call}.
	
	\subsection{Speech Recognition Models}
	
	All the models take 80-channel log-mel filterbank features as input and are trained end-to-end with the Connectionist Temporal Classification (CTC) criterion~\cite{graves2006connectionist}. The target vocabulary is a wordpiece model~\cite{schuster2012japanese} with size 10,000. Models are all trained in the \textit{wav2letter++} framework~\cite{pratap2018wav2letter} using either the \librispeech{} or FB Videos.
	\\
	\textbf{Model trained on full LS:}
	We use the Transformer model from \cite{synnaeve2019end} that works best on the \librispeech{}\footnote{
		\url{github.com/facebookresearch/wav2letter/tree/master/recipes/models/sota/2019}.}. Specifically, there are 6 layers of 1-D convolutions with kernel width 3 as front-end followed by 24 4-head Transformer blocks with self-attention dimension 1024. The 2nd, 4th and the last convolutions in the front-end have stride 2, so the overall sub-sampling of the model is 8.
	\\
	\textbf{Model trained on LS 100h:} 
	We use a similar model as for full LS. In order to obtain better performance with a small amount of training data, we use 24 4-head Transformer blocks with self-attention dimension 768 in the middle.
	\\
	\textbf{Model trained on FB Videos:} 
	The model is mainly built upon Time-Depth Separable Convolution (TDS)~\cite{hannun2019TDS} blocks. It is composed of one 2-D convolution layer and two fully-connected layers with ReLU, LayerNorm and residual connections in between. Specifically, the model has $4$ groups of TDS blocks with a 1-D convolutions at the beginning of each group as transitions. Similarly, the first 3 convolutions have stride 2 so as to reach the same sub-sampling rate of 8. There are $2$, $2$, $5$, and $8$ TDS blocks in each group, containing $16$, $16$, $24$, and $32$ channels, respectively. Following \cite{synnaeve2019end}, we also apply a channel increasing factor $F=2$ in each TDS block.
	\\
	\textbf{Language model:}
	A language model (LM) is integrated in the beam-search decoder to generate final transcriptions together with the acoustic models. In our experiments, we use 4-gram LMs trained with KenLM toolkit~\cite{heafield2011kenlm}. The LM used for the FB Videos is trained on the transcriptions, while the one for the \librispeech{} is trained on its official LM corpus excluding books containing the transcriptions of \librilight{} audios. The latter LM is prepared in \cite{synnaeve2019end}.
	
	After the acoustic models converged on the labeled data, we tune the beam-search decoder parameters on the validation set. 
	Specifically, the decoder consumes the posterior from the acoustic model, and runs a beam search through with LM to generate the best path, with beam size 300 for the small Transformer (768) and 500 for the large Transformer (1024).
	
	\begin{table}[t]
		\centering
		\caption{Number of parameters for each model architecture.}
		\begin{tabular}{llr}
			\toprule
			Task & Model & \# Parameters \\
			\midrule
			\multirow{3}{*}{ST} & LSTM & 13.5M \\
			& \textsc{VggT} & 260.0M \\
			& \textsc{VggTLarge} & 435.0M \\
			\cmidrule(lr){1-3}
			\multirow{3}{*}{ASR} & Transformer 1024 & 339.9M \\
			& Transformer 768 & 204.7M \\
			& TDS & 292.0M \\
			\cmidrule(lr){1-3}
			\multirow{4}{*}{MT} & En-Es FB Video & 320.1M \\
			& En-Fr FB Video & 300.6M \\
			& En-De \cite{ott-etal-2018-scaling} & 209.9M \\
			& En-Fr \cite{ott-etal-2018-scaling} & 221.9M \\
			\bottomrule
		\end{tabular}
		\label{tab:model_params}
	\end{table}
	
	\subsection{Machine Translation Models}
	
	All MT models use a Transformer~\cite{vaswani2017attention} architecture. The model for FB Videos uses 6 encoder layers, encoder embedding dimension of 1024, encoder feed-forward network (FFN) dimension of 2048, 16 attention heads, 2 decoder layers, decoder embedding dimension of 512, decoder FFN dimension of 1024, dropout of 0.2, and label smoothing of 0.1. A bottleneck linear layer with dimension 128 is inserted prior to the softmax over the target vocabulary and the decoder is an average attention network~\cite{zhang2018accelerating}. 
	The final model is an ensemble of size 3 obtained from 3 training runs started with different random seeds, where for each training runs, the last 10 checkpoints are averaged. The original Transformer optimizer settings are used, with an initial learning rate of 0.0007 and an effective batch size of \num{64000} tokens. The models are trained on 100M sentence pairs from the web, news and social media domain until convergence. In inference, hypotheses are generated with beam search with beam size 2. \textsc{LibriSpeech} transcripts and \textsc{LibriLight} automatic transcripts are translated with pre-trained\footnote{\url{github.com/pytorch/fairseq/tree/master/examples/translation}} English-French and English-German models~\cite{ott-etal-2018-scaling}, with beam size 5.
	Model sizes are summarized in \autoref{tab:model_params}.

	\section{Results}
	
	
	In this section, we study under which conditions pseudo-labels generated by the cascade model can benefit the ST model. 
	
	\subsection{Adding \textsc{LibriLight} pseudo-labels}
	\label{sec:low_resource}
	
\begin{figure}[h!]
		\centering
		\begin{tikzpicture}
		\definecolor{darkgreen}{RGB}{0, 100, 0}
		rgb(0,100,0)
		\begin{groupplot}
		[group style={group size= 1 by 1, horizontal sep=0.5cm, vertical sep=1.2 cm}, width=0.85\columnwidth, legend cell align={left}, height = 4.8cm, legend style={font=\notsotiny}, label style={font=\scriptsize}, tick label style={font=\scriptsize}]
		\nextgroupplot[
		ylabel=BLEU,
		y label style={yshift=-15pt},
		ymin=12, 
		ymax=20,
		xlabel=Amount of \textsc{LibriLight} data (hours),
		xmode=log,
		xmin=30, 
		xmax=60000,
		legend style={legend pos=south west},
		xminorticks=false
		]
		\addplot[red, densely dashed] coordinates {(1,16.3) (60000,16.3)};
		\addlegendentry{MuST-C Baseline};
		\addplot[cyan, mark=*] table [x index=0,y index=1, dashed, col sep=comma] {mustc_ll.csv};
		\addlegendentry{MuST-C + LL};
		\addplot[darkgreen, densely dashed] coordinates {(1,19.2) (60000,19.2)};
		\addlegendentry{MuST-C + LS Baseline};
		\addplot[blue, mark=*] table [x index=0,y index=1, dashed, col sep=comma] {mustc_lsll.csv};
		\addlegendentry{MuST-C + LS + LL};
		\end{groupplot}
		\end{tikzpicture}
		\caption{Results obtained with additional \textsc{LibriLight} (LL) data and the LSTM architecture on the En-De MuST-C dev set.}
		\label{fig:mustc_ll_lsll}
	\end{figure}
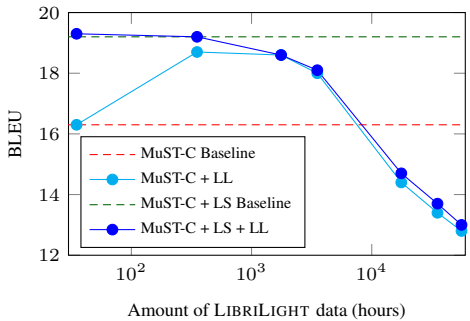

	We first study the effect of simply adding pseudo-labels to the baseline training
	data and retraining with the LSTM model, where the baseline is either lower resource (MuST-C) or higher resource (MuST-C + LS).
	In \autoref{fig:mustc_ll_lsll}, with this simple method, the low resource baseline can be improved by up to 2.4 BLEU on the En-De MuST-C dev set. Above a certain amount, adding unlabeled data degrades performance.
	However, the higher resource baseline is not improved by simply adding unlabeled data. Next, our focus is on improving the higher resource baseline and on leveraging larger amounts of unlabeled data.
	
	\subsection{Improving a High-Resource Baseline}
	\label{sec:finetuning_transformer}
	
\begin{figure}[h!]
		\centering
		\begin{tikzpicture}
		\definecolor{darkgreen}{RGB}{0, 100, 0}
		\begin{groupplot}
		[group style={group size= 1 by 1, horizontal sep=0.5cm, vertical sep=1.2 cm}, width=0.85\columnwidth, legend cell align={left}, height = 4.8cm, legend style={font=\notsotiny}, label style={font=\scriptsize}, tick label style={font=\scriptsize}]
		\nextgroupplot[
		ylabel=BLEU,
		y label style={yshift=-15pt},
		ymin=16, 
		ymax=26,
		xlabel=Amount of \textsc{LibriLight} data (hours),
		xmode=log,
		xmin=30, 
		xmax=60000,
		legend style={legend pos=north west},
		xminorticks=false
		]
		\addplot[red, dashed] coordinates {(1,19.2) (60000,19.2)};
		\addlegendentry{Baseline};
		\addplot[blue, mark=*] table [x index=0,y index=1, dashed, col sep=comma] {mustc_lsll_finetune.csv};
		\addlegendentry{+ LL, LSTM};
		\addplot[darkgreen, mark=*] table [x index=0,y index=1, dashed, col sep=comma] {mustc_lsll_transformer_finetune.csv};
		\addlegendentry{+ LL, \textsc{VggT}};
		\end{groupplot}
		\end{tikzpicture}
		\caption{Improving the higher resource MuST-C + LS baseline on the En-De MuST-C dev set by fine-tuning the LSTM model and training and fine-tuning a larger architecture, \textsc{VggT}.}
		\label{fig:mustc_lsll_finetune_transformer}
	\end{figure}
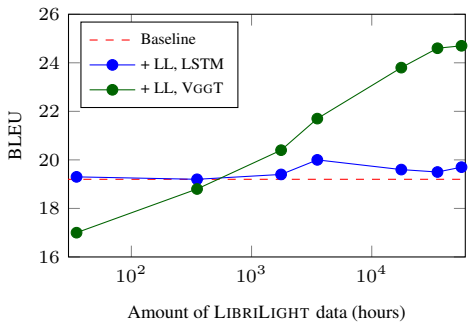

	In \autoref{fig:mustc_lsll_finetune_transformer}, the higher resource baseline can be improved upon by fine-tuning the LSTM model on the baseline data (MuST-C + LS) and	by training (and fine-tuning) \textsc{VggT}. Fine-tuning simply consists in loading the latest checkpoint from the initial training phase and continuing training on the baseline training data (without resetting the optimizer parameters). We obtain up to 24.7 BLEU, i.e., a 5.5 BLEU gain over the high resource baseline. Note that training the \textsc{VggT} model on the baseline data does not converge and yields only 3.7 BLEU.
	%
	
	\subsection{Scaling Model Size}
	\label{sec:scaling_model_size}
	
	In \autoref{sec:finetuning_transformer}, substantial improvements over the baseline were obtained by training a larger model. In \autoref{tab:scaling_up}, the capacity of the model is further increased in order to verify to what extent pseudo-labels can benefit training. When adding \SI{17607}{\hour} of unlabeled data, \textsc{VggT} and \textsc{VggTLarge}
	obtain similar performance but with \SI{35217}{\hour} of additional data, \textsc{VggTLarge} obtains 1 BLEU improvement on the En-De MuST-C dev set.
	%
	%
	\begin{table}[]
		\centering
		\caption{Results obtained on the En-De MuST-C dev set when increasing the model size. Results are reported after fine-tuning.}
		\begin{tabular}{lrr}
			\toprule
			Data & \textsc{VggT} & \text{VggTLarge} \\
			\midrule
			MuST-C + LS + \num{17607}h LL & 23.8 & 23.7 \\
			MuST-C + LS + \num{35217}h LL & 24.6 & 25.6 \\
			\bottomrule
		\end{tabular}
		\label{tab:scaling_up}
	\end{table}
	
	\subsection{Main Results}
	
\begin{figure}[h!]
		\centering
		\begin{tikzpicture}
		\begin{groupplot}
		[group style = {group size = 2 by 1, horizontal sep = 0.71cm}, width =4.7cm, height =4.0cm],
		width=3.5cm]
		\nextgroupplot[
		ylabel=BLEU,
		y label style={yshift=-0.53cm},
		ymin=16, 
		ymax=21,
		xmode=log,
		xmin=400,
		xmax=100000,
		title={\scriptsize En-Fr},
		xminorticks=false,
		label style={font=\scriptsize},
		tick label style={font=\scriptsize}
		]
		\addplot[red, dashed] coordinates {(1,19.262827480058) (100000,19.262827480058)};
		\addplot[cyan, mark=*] table [x index=0,y index=1, dashed, col sep=comma] {video_ll_en_fr.csv};
		\addplot[blue, mark=*] table [x index=0,y index=1, dashed, col sep=comma] {video_ll_finetune_en_fr.csv};
		\nextgroupplot[
		ymin=20, 
		ymax=26,
		xlabel=Amount of unlabeled data (hours),
		x label style={xshift=-45pt},
		xmode=log,
		xmin=400,
		xmax=100000,
		title={\scriptsize En-Es},
		legend style={font=\notsotiny, at={(0.65,-0.44)},legend columns=3},
		xminorticks=false,
		label style={font=\scriptsize},
		tick label style={font=\scriptsize}
		]
		\addplot[red, dashed] coordinates {(1,24.237125355315) (100000,24.237125355315)};
		\addlegendentry{Baseline};
		\addplot[cyan, mark=*] table [x index=0,y index=1, dashed, col sep=comma] {video_ll_en_es.csv};
		\addlegendentry{+ LL};
		\addplot[blue, mark=*] table [x index=0,y index=1, dashed, col sep=comma] {video_ll_finetune_en_es.csv};
		\addlegendentry{+ LL fine-tuned};
		\end{groupplot}
		\end{tikzpicture}
		\caption{Effectiveneness of pseudo-labeling on FB Videos for En-Fr and En-Es. Results are reported on the dev set.}
		\label{fig:internal_exps}
	\end{figure}
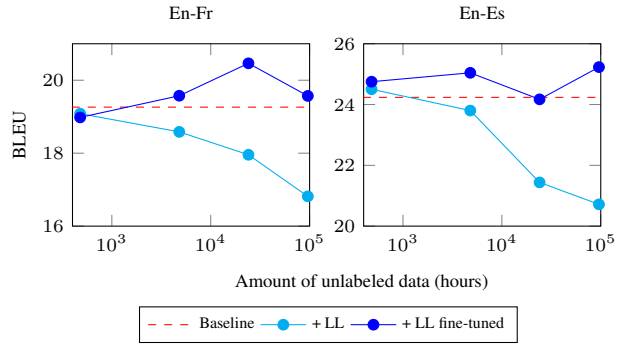

	We now validate our findings on three languages and two domains.
	In the En-Fr and En-Es FB Video setting, different amounts of unlabeled data are added to the baseline data, then \textsc{VggT} is retrained and fine-tuned. \autoref{fig:internal_exps} confirms earlier conclusions that fine-tuning is necessary to obtain improvements over a strong high-resource baseline. We obtain up to 1.2 and 1.0 BLEU gains on the En-Fr and En-Es dev sets, respectively.
	Final results on the test sets on three language pairs and two domains are summarized in \autoref{tab:main_results}. On the En-Fr MuST-C tst-COMMON dataset, we obtain 8.3 BLEU improvements over the strong MuST-C + LS baseline and improve state-of-the-art from \cite{9054759} by 0.45 BLEU. On the En-De MusT-C tst-COMMON dataset, we obtain 5.7 BLEU improvements over the MusT-C + LS baseline and improve the state-of-the-art by 3.1 BLEU. Finally, we verify that our method works with a very large-scale FB video baseline by obtaining 1.3 and 1.4 BLEU gains on the FB Video En-Fr and En-Es test sets, respectively.
	%
	\begin{table}[]
		\centering
		\caption{Leveraging unlabeled audio on 3 language pairs and 2 domains. Results are reported on the MuST-C tst-COMMON sets and the FB Video test sets.}
		\resizebox{\columnwidth}{!}{%
			\begin{tabular}{llrr}
				\toprule
				Language & Data & Model & BLEU \\
				\midrule
				\multirow{6}{*}{En-Fr} & MuST-C & \multirow{2}{*}{LSTM} & 24.8 \\
				& MuST-C + LS &  & 26.2 \\
				\cmidrule(lr){2-4}
				& MuST-C + LS & \multirow{2}{*}{\textsc{VggT}} & 23.9 \\
				& + \num{35217}h LL + fine-tuning & & \textbf{34.5} \\
				\cmidrule(lr){2-4}
				\cmidrule(lr){2-4}
				& State-of-the-art baseline~\cite{9054759} & & 34.05 \\
				\midrule
				\multirow{6}{*}{En-De}  & MuST-C & \multirow{2}{*}{LSTM} & 15.6 \\
				& MuST-C + LS &  & 19.5 \\
				\cmidrule(lr){2-4}
				& MuST-C + LS & \multirow{2}{*}{\textsc{VggT}} & 3.5 \\
				& + \num{35217}h LL + fine-tuning & & 24.8 \\
				\cmidrule(lr){2-4}
				& + \num{35217}h LL + fine-tuning & \textsc{VggTLarge} & \textbf{25.2} \\
				\cmidrule(lr){2-4}
				& State-of-the-art baseline~\cite{9054759} & & 22.11 \\
				\midrule
				\multirow{2}{*}{\shortstack[l]{En-Fr\\(FB Videos)}} & baseline & \multirow{2}{*}{\textsc{VggT}} & \metric{20.25255581} \\
				& + 96k h unlabeled + fine-tuning & & \textbf{21.6} \\
				\midrule
				\multirow{2}{*}{\shortstack[l]{En-Es\\(FB Videos)}} & baseline & \multirow{2}{*}{\textsc{VggT}} & \metric{18.45664391} \\
				& + 96k h unlabeled + fine-tuning & & \textbf{19.9} \\
				\bottomrule
			\end{tabular}
		}
		\label{tab:main_results}
	\end{table}
	
	\section{Ablation Studies}
	
	\subsection{Pseudo-Labeling vs.\ ASR Encoder Pre-training}
	\label{sec:selftraining_vs_pretraining}
	
	In \autoref{sec:finetuning_transformer}, pseudo-labels enable training much larger architectures that are otherwise difficult to train. In this section, we investigate whether this regularization effect is simply due to better pre-training of the encoder. To verify this, the \textsc{VggT} architecture is first trained on the ASR task exactly as in the ST task, then the encoder is initialize with the parameters obtained in ASR training and the same model is trained on the ST task. Data conditions include the high-resource baseline and the baseline augmented with \num{3523} hours and \num{17607} hours from \textsc{LibriLight}.
	Results are reported in \autoref{tab:selftraining_vs_pretraining}. The word error rate (WER) obtained on the ASR task is also reported for the encoder pre-training method (the higher WER obtained with \num{17607}h of data is simply due to the large amount of weakly supervised data but this setting still benefits the ST task). Except in the baseline setting, pseudo-labeling outperforms encoder pre-training, by up to 2.9 BLEU. This highlights the importance of both encoder and decoder pre-training.
	\begin{table}[]
		\centering
		\caption{Comparing self-training and pre-training the encoder on the ASR task. Results are reported on the En-De MuST-C dev set after fine-tuning.}
		\begin{tabular}{lrr}
			\toprule
			\multirow{2}{*}{Data} & Encoder Pre-training & Pseudo-Labeling \\
			& BLEU (WER) & BLEU \\
			\midrule
			MuST-C + LS & 19.8 (25.2) & 3.7 \\
			+ \SI{3523}{\hour} LL & 20.8 (22.7) & 21.7 \\
			+ \SI{17607}{\hour} LL & 21.0 (40.3) & 23.9 \\ 
			\bottomrule
		\end{tabular}
		\label{tab:selftraining_vs_pretraining}
	\end{table}
	%
	
	\subsection{Quality of Pseudo-Labels}
	\label{sec:quality}
	
\begin{figure}[h!]
		\centering
		\begin{tikzpicture}
		\begin{groupplot}
		[group style={group size= 1 by 1, horizontal sep=0.5cm, vertical sep=1.2 cm}, width=0.85\columnwidth, legend cell align={left}, height = 4.2cm, legend style={font=\notsotiny}, label style={font=\scriptsize}, tick label style={font=\scriptsize}]
		\nextgroupplot[
		ylabel=BLEU,
		y label style={yshift=-15pt},
		ymin=12, 
		ymax=20,
		xlabel=Amount of \textsc{LibriLight} data (hours),
		xmode=log,
		xmin=30, 
		xmax=60000,
		legend style={legend pos=south west},
		xminorticks=false
		]
		\addplot[red, dashed] coordinates {(1,16.3) (60000,16.3)};
		\addlegendentry{MuST-C Baseline};
		\addplot[cyan, mark=*] table [x index=0,y index=1, dashed, col sep=comma] {mustc_ll.csv};
		\addlegendentry{MuST-C + LL};
		\addplot[blue, mark=*] table [x index=0,y index=1, dashed, col sep=comma] {mustc_ll_lq.csv};
		\addlegendentry{MuST-C + LL LQ};
		\end{groupplot}
		\end{tikzpicture}
		\caption{Effect of using lower quality (LQ) pseudo-labels. Results are reported on the En-De MuST-C dev set.}
		\label{fig:quality_ablation}
	\end{figure}
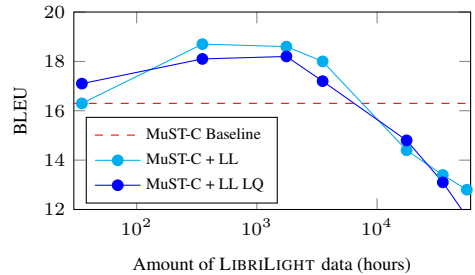

	The effect of the quality of pseudo-labels is now investigated. Automatic transcripts are generated either with
	the ASR model trained on the full \librispeech{} or on a 100 hour subset, then translated with the same translation system. The two models obtain 7.3 and 27.7 WER on the \textsc{LibriSpeech} \devother{} set. The LSTM speech translation model is then retrained on both types of labels with different data amounts. As expected, \autoref{fig:quality_ablation} shows that in the majority of data conditions, the BLEU score increases with higher quality labels.
	
	\subsection{Self-Training}
	\label{sec:selftraining}
	
	So far, pseudo-labels have been generated via a cascade model.
	In this section, two end-to-end models are considered for pseudo-label generation. The pure self-training scenario where the LSTM end-to-end model has only been trained on the supervised training data is first considered. Pseudo-labels are also generated with the \textsc{VggT} model trained on MuST-C, LS and \num{17607}h of cascade-generated \textsc{LibriLight} pseudo-labels. We contrast pseudo-label generation by the cascade model and these two models in \autoref{tab:self_training}. First, all pseudo-labeling methods improve upon the baseline, even the pure self-training method. The weakest pseudo-labeling method is the pure self-training method that does not use extra information in the process. In the lower resource baseline setting, the cascade and \textsc{VggT} obtain equivalent performance, the cascade having a slight advantage of 0.2 BLEU. In the higher resource setting, the \textsc{VggT} end-to-end pseudo-labeling method
	outperforms the cascade pseudo-labeling by 0.7 BLEU. We conclude that cascade pseudo-labeling can to bootstrap the pseudo-labeling process, then end-to-end ST can be relied on for pseudo-labeling in subsequent iterations.
	\begin{table}[]
		\centering
		\caption{Pseudo-labeling with end-to-end speech translation. Results are reported on the En-De MuST-C dev set, after fine-tuning.}
		\begin{tabular}{lllr}
			\toprule
			\multirow{2}{*}{Data} & Pseudo-Labeling & \multirow{2}{*}{Model} & \multirow{2}{*}{BLEU} \\
			& Model & & \\
			\midrule
			MuST-C & N/A & LSTM & 16.3 \\
			\cmidrule(lr){1-4}
			\multirow{3}{*}{+ \num{3523}h LL} & Cascade & \multirow{3}{*}{LSTM} & 20.8 \\
			& LSTM & & 18.5 \\
			& \textsc{VggT} & & 20.6 \\
			\midrule
			MuST-C + LS & N/A & LSTM & 19.2 \\
			\cmidrule(lr){1-4}
			\multirow{3}{*}{+ \num{17607}h LL} & Cascade & \multirow{3}{*}{\textsc{VggT}} & 23.8 \\
			& LSTM &  & 20.7 \\
			& \textsc{VggT} &  & \textbf{24.5} \\
			\bottomrule
		\end{tabular}
		\label{tab:self_training}
	\end{table}

	\section{Conclusion}
	
	We have shown the effectiveness of pseudo-labels for end-to-end ST in low- and high-resource data conditions, across two domains and 3 language pairs. In the high-resource setting, fine-tuning and larger architectures were found to be critical for obtaining improvements over the baseline. Larger amounts of pseudo-labels allow to increase the model size further. By doing so, we obtained state-of-the-art results on the MuST-C English-French and English-German datasets. Our approach was shown empirically to be more effective than encoder pre-training, highlighting the importance of pre-training the decoder. Finally, pseudo-labeling may be further simplified by utilizing end-to-end ST systems instead of a cascade system.
	
	\flushend

	
	

	\newpage
	\bibliographystyle{IEEEtran}
	
	\bibliography{selftraining}

\begin{thebibliography}{10}
\providecommand{\url}[1]{#1}
\csname url@samestyle\endcsname
\providecommand{\newblock}{\relax}
\providecommand{\bibinfo}[2]{#2}
\providecommand{\BIBentrySTDinterwordspacing}{\spaceskip=0pt\relax}
\providecommand{\BIBentryALTinterwordstretchfactor}{4}
\providecommand{\BIBentryALTinterwordspacing}{\spaceskip=\fontdimen2\font plus
\BIBentryALTinterwordstretchfactor\fontdimen3\font minus
  \fontdimen4\font\relax}
\providecommand{\BIBforeignlanguage}[2]{{%
\expandafter\ifx\csname l@#1\endcsname\relax
\typeout{** WARNING: IEEEtran.bst: No hyphenation pattern has been}%
\typeout{** loaded for the language `#1'. Using the pattern for}%
\typeout{** the default language instead.}%
\else
\language=\csname l@#1\endcsname
\fi
#2}}
\providecommand{\BIBdecl}{\relax}
\BIBdecl

\bibitem{jia2019leveraging}
Y.~Jia, M.~Johnson, W.~Macherey, R.~J. Weiss, Y.~Cao, C.-C. Chiu, N.~Ari,
  S.~Laurenzo, and Y.~Wu, ``Leveraging weakly supervised data to improve
  end-to-end speech-to-text translation,'' in \emph{ICASSP 2019-2019 IEEE
  International Conference on Acoustics, Speech and Signal Processing
  (ICASSP)}.\hskip 1em plus 0.5em minus 0.4em\relax IEEE, 2019, pp. 7180--7184.

\bibitem{pino2019harnessing}
J.~Pino, L.~Puzon, J.~Gu, X.~Ma, A.~D. McCarthy, and D.~Gopinath, ``Harnessing
  indirect training data for end-to-end automatic speech translation: Tricks of
  the trade,'' in \emph{Proceedings of the 16th International Workshop on
  Spoken Language Translation (IWSLT)}, 2019.

\bibitem{Weiss2017}
\BIBentryALTinterwordspacing
R.~J. Weiss, J.~Chorowski, N.~Jaitly, Y.~Wu, and Z.~Chen,
  ``Sequence-to-sequence models can directly translate foreign speech,'' in
  \emph{Proc. Interspeech 2017}, 2017, pp. 2625--2629. [Online]. Available:
  \url{http://dx.doi.org/10.21437/Interspeech.2017-503}
\BIBentrySTDinterwordspacing

\bibitem{berard2018end}
A.~B{\'e}rard, L.~Besacier, A.~C. Kocabiyikoglu, and O.~Pietquin, ``End-to-end
  automatic speech translation of audiobooks,'' in \emph{2018 IEEE
  International Conference on Acoustics, Speech and Signal Processing
  (ICASSP)}.\hskip 1em plus 0.5em minus 0.4em\relax IEEE, 2018, pp. 6224--6228.

\bibitem{bansal-etal-2019-pre}
\BIBentryALTinterwordspacing
S.~Bansal, H.~Kamper, K.~Livescu, A.~Lopez, and S.~Goldwater, ``Pre-training on
  high-resource speech recognition improves low-resource speech-to-text
  translation,'' in \emph{Proceedings of the 2019 Conference of the North
  {A}merican Chapter of the Association for Computational Linguistics: Human
  Language Technologies, Volume 1 (Long and Short Papers)}.\hskip 1em plus
  0.5em minus 0.4em\relax Minneapolis, Minnesota: Association for Computational
  Linguistics, Jun. 2019, pp. 58--68. [Online]. Available:
  \url{https://www.aclweb.org/anthology/N19-1006}
\BIBentrySTDinterwordspacing

\bibitem{9053847}
M.~C. {Stoian}, S.~{Bansal}, and S.~{Goldwater}, ``Analyzing {ASR} pretraining
  for low-resource speech-to-text translation,'' in \emph{ICASSP 2020 - 2020
  IEEE International Conference on Acoustics, Speech and Signal Processing
  (ICASSP)}, 2020, pp. 7909--7913.

\bibitem{9004003}
M.~A. {Di Gangi}, M.~{Negri}, and M.~{Turchi}, ``One-to-many multilingual
  end-to-end speech translation,'' in \emph{2019 IEEE Automatic Speech
  Recognition and Understanding Workshop (ASRU)}, 2019, pp. 585--592.

\bibitem{9003832}
H.~{Inaguma}, K.~{Duh}, T.~{Kawahara}, and S.~{Watanabe}, ``Multilingual
  end-to-end speech translation,'' in \emph{2019 IEEE Automatic Speech
  Recognition and Understanding Workshop (ASRU)}, 2019, pp. 570--577.

\bibitem{wang2020covost}
C.~Wang, J.~Pino, A.~Wu, and J.~Gu, ``Covost: A diverse multilingual
  speech-to-text translation corpus,'' \emph{arXiv preprint arXiv:2002.01320},
  2020.

\bibitem{scudder1965probability}
H.~Scudder, ``Probability of error of some adaptive pattern-recognition
  machines,'' \emph{IEEE Transactions on Information Theory}, vol.~11, no.~3,
  pp. 363--371, 1965.

\bibitem{He2020Revisiting}
\BIBentryALTinterwordspacing
J.~He, J.~Gu, J.~Shen, and M.~Ranzato, ``Revisiting self-training for neural
  sequence generation,'' in \emph{International Conference on Learning
  Representations}, 2020. [Online]. Available:
  \url{https://openreview.net/forum?id=SJgdnAVKDH}
\BIBentrySTDinterwordspacing

\bibitem{synnaeve2019end}
G.~Synnaeve, Q.~Xu, J.~Kahn, T.~Likhomanenko, E.~Grave, V.~Pratap, A.~Sriram,
  V.~Liptchinsky, and R.~Collobert, ``End-to-end {ASR}: from supervised to
  semi-supervised learning with modern architectures,'' in \emph{Proceedings of
  the ICML 2020 Workshop on Self-supervision in Audio and Speech}, 2020.

\bibitem{kahn2019self}
J.~Kahn, A.~Lee, and A.~Hannun, ``Self-training for end-to-end speech
  recognition,'' \emph{arXiv preprint arXiv:1909.09116}, 2019.

\bibitem{librilight}
J.~{Kahn}, M.~{Rivière}, W.~{Zheng}, E.~{Kharitonov}, Q.~{Xu}, P.~E.
  {Mazaré}, J.~{Karadayi}, V.~{Liptchinsky}, R.~{Collobert}, C.~{Fuegen},
  T.~{Likhomanenko}, G.~{Synnaeve}, A.~{Joulin}, A.~{Mohamed}, and E.~{Dupoux},
  ``Libri-light: A benchmark for {ASR} with limited or no supervision,'' in
  \emph{ICASSP 2020 - 2020 IEEE International Conference on Acoustics, Speech
  and Signal Processing (ICASSP)}, 2020, pp. 7669--7673.

\bibitem{yalniz2019billion}
I.~Z. Yalniz, H.~J{\'e}gou, K.~Chen, M.~Paluri, and D.~Mahajan, ``Billion-scale
  semi-supervised learning for image classification,'' \emph{arXiv preprint
  arXiv:1905.00546}, 2019.

\bibitem{9054759}
S.~{Indurthi}, H.~{Han}, N.~K. {Lakumarapu}, B.~{Lee}, I.~{Chung}, S.~{Kim},
  and C.~{Kim}, ``End-end speech-to-text translation with modality agnostic
  meta-learning,'' in \emph{ICASSP 2020 - 2020 IEEE International Conference on
  Acoustics, Speech and Signal Processing (ICASSP)}, 2020, pp. 7904--7908.

\bibitem{di-gangi-etal-2019-must}
\BIBentryALTinterwordspacing
M.~A. Di~Gangi, R.~Cattoni, L.~Bentivogli, M.~Negri, and M.~Turchi,
  ``{M}u{ST}-{C}: a {M}ultilingual {S}peech {T}ranslation {C}orpus,'' in
  \emph{Proceedings of the 2019 Conference of the North {A}merican Chapter of
  the Association for Computational Linguistics: Human Language Technologies,
  Volume 1 (Long and Short Papers)}.\hskip 1em plus 0.5em minus 0.4em\relax
  Minneapolis, Minnesota: Association for Computational Linguistics, Jun. 2019,
  pp. 2012--2017. [Online]. Available:
  \url{https://www.aclweb.org/anthology/N19-1202}
\BIBentrySTDinterwordspacing

\bibitem{panayotov2015librispeech}
V.~Panayotov, G.~Chen, D.~Povey, and S.~Khudanpur, ``Librispeech: {A}n {ASR}
  corpus based on public domain audio books,'' in \emph{2015 IEEE International
  Conference on Acoustics, Speech and Signal Processing (ICASSP)}.\hskip 1em
  plus 0.5em minus 0.4em\relax IEEE, 2015, pp. 5206--5210.

\bibitem{Kudo2018SentencePieceAS}
T.~Kudo and J.~Richardson, ``Sentencepiece: A simple and language independent
  subword tokenizer and detokenizer for neural text processing,'' in
  \emph{EMNLP}, 2018.

\bibitem{vaswani2017attention}
A.~Vaswani, N.~Shazeer, N.~Parmar, J.~Uszkoreit, L.~Jones, A.~N. Gomez,
  {\L}.~Kaiser, and I.~Polosukhin, ``Attention is all you need,'' in
  \emph{Advances in neural information processing systems}, 2017, pp.
  5998--6008.

\bibitem{mohamed2019transformers}
A.~Mohamed, D.~Okhonko, and L.~Zettlemoyer, ``Transformers with convolutional
  context for {ASR},'' \emph{arXiv preprint arXiv:1904.11660}, 2019.

\bibitem{Kingma2015AdamAM}
D.~P. Kingma and J.~Ba, ``Adam: A method for stochastic optimization,'' in
  \emph{ICLR}, vol. abs/1412.6980, 2015.

\bibitem{post-2018-call}
\BIBentryALTinterwordspacing
M.~Post, ``A call for clarity in reporting {BLEU} scores,'' in
  \emph{Proceedings of the Third Conference on Machine Translation: Research
  Papers}.\hskip 1em plus 0.5em minus 0.4em\relax Brussels, Belgium:
  Association for Computational Linguistics, Oct. 2018, pp. 186--191. [Online].
  Available: \url{https://www.aclweb.org/anthology/W18-6319}
\BIBentrySTDinterwordspacing

\bibitem{graves2006connectionist}
A.~Graves, S.~Fern{\'a}ndez, F.~Gomez, and J.~Schmidhuber, ``Connectionist
  temporal classification: {L}abelling unsegmented sequence data with recurrent
  neural networks,'' in \emph{Proceedings of the 23rd international conference
  on Machine learning}, 2006, pp. 369--376.

\bibitem{schuster2012japanese}
M.~Schuster and K.~Nakajima, ``Japanese and {K}orean voice search,'' in
  \emph{2012 IEEE International Conference on Acoustics, Speech and Signal
  Processing (ICASSP)}.\hskip 1em plus 0.5em minus 0.4em\relax IEEE, 2012, pp.
  5149--5152.

\bibitem{pratap2018wav2letter}
V.~Pratap, A.~Hannun, Q.~Xu \emph{et~al.}, ``wav2letter++: The fastest
  open-source speech recognition system,'' \emph{arXiv preprint
  arXiv:1812.07625}, 2018.

\bibitem{hannun2019TDS}
A.~Hannun, A.~Lee, Q.~Xu, and R.~Collobert, ``Sequence-to-sequence speech
  recognition with time-depth separable convolutions,'' \emph{Interspeech
  2019}, Sep 2019.

\bibitem{heafield2011kenlm}
\BIBentryALTinterwordspacing
K.~Heafield, ``{K}en{LM}: Faster and smaller language model queries,'' in
  \emph{Proceedings of the Sixth Workshop on Statistical Machine
  Translation}.\hskip 1em plus 0.5em minus 0.4em\relax Edinburgh, Scotland:
  Association for Computational Linguistics, Jul. 2011, pp. 187--197. [Online].
  Available: \url{https://www.aclweb.org/anthology/W11-2123}
\BIBentrySTDinterwordspacing

\bibitem{ott-etal-2018-scaling}
\BIBentryALTinterwordspacing
M.~Ott, S.~Edunov, D.~Grangier, and M.~Auli, ``Scaling neural machine
  translation,'' in \emph{Proceedings of the Third Conference on Machine
  Translation: Research Papers}.\hskip 1em plus 0.5em minus 0.4em\relax
  Brussels, Belgium: Association for Computational Linguistics, Oct. 2018, pp.
  1--9. [Online]. Available: \url{https://www.aclweb.org/anthology/W18-6301}
\BIBentrySTDinterwordspacing

\bibitem{zhang2018accelerating}
B.~Zhang, D.~Xiong, and J.~Su, ``Accelerating neural transformer via an average
  attention network,'' \emph{arXiv preprint arXiv:1805.00631}, 2018.

\end{thebibliography}
	
	
\end{document}